\documentclass[letterpaper, 10 pt, conference]{ieeeconf}  
\IEEEoverridecommandlockouts                              
\pdfoutput=1
\usepackage{graphicx}
\usepackage{epsfig} 
\usepackage{amsmath} 

\usepackage{amssymb}  
\usepackage{amsthm}
\usepackage[mathcal]{euscript}
\usepackage{mathrsfs}
\usepackage{bm}

\usepackage{color}
\usepackage{lipsum}
\usepackage[colorlinks=true,linkcolor=black,anchorcolor=black,citecolor=black,filecolor=black,menucolor=black,runcolor=black,urlcolor=cyan]{hyperref}
\usepackage{etoolbox}
\usepackage[table,xcdraw,dvipsnames]{xcolor}
\usepackage{multirow}
\usepackage{mathtools}
\usepackage{nopageno}
\usepackage{algorithm}
\usepackage{algpseudocode}
\usepackage{xspace}

\hypersetup{
    colorlinks=true,
    linkcolor=blue,
    filecolor=magenta,      
    urlcolor=cyan,
    pdftitle={CROWS},
    pdfpagemode=FullScreen,
    }

\usepackage{outlines}
\let\labelindent\relax
\usepackage{enumitem}
\setenumerate[1]{label=\Roman*.}
\setenumerate[2]{label=\Alph*.}
\setenumerate[3]{label=\arabic*.}
\setenumerate[4]{label=\roman*.}

\usepackage{marginnote}

\usepackage[dvipsnames]{xcolor}
\usepackage{changes}

\usepackage{pifont}

\usepackage{booktabs}

\makeatletter
\patchcmd{\@makecaption}
  {\scshape}
  {}
  {}
  {}
\makeatother

\usepackage[
    style=ieee,
    doi=false,
    isbn=false,
    url=false,
    eprint=false,
    backend=biber,
    natbib=true
    ]{biblatex}
\bibliography{references}

\title{\LARGE \bf Conformalized Reachable Sets for Obstacle Avoidance With Spheres}

\author{Yongseok Kwon$^{1}$, Jonathan Michaux$^{1}$, Seth Isaacson$^{1}$, Bohao Zhang$^{1}$, Matthew Ejakov$^{1}$,\\ Katherine A. Skinner$^{1}$ and Ram Vasudevan$^{1}$
\thanks{This work is supported by the National Science Foundation Career Award \#1751093 and by the Air Force Office of Scientific Research under award 23-S15.}
\thanks{$^{1}$Robotics Institute, University of Michigan, Ann Arbor, MI $\langle$\texttt{kwonys, jmichaux, sethgi, jimzhang, ejakovm, kskin, ramv}$\rangle$\texttt{@umich.edu}.}}

\begin{document}

\newcommand{\red}[1]{{\color{red} #1}}

\providecommand{\sidenote}[1]{\textcolor{red}{\marginnote{#1}}}

\providecommand{\Review}[3]{\textcolor{red}{{\footnotesize \protect\marginnote{#3}}\sout{#1}\red{{#2}}}}

\providecommand{\ReviewR}[3]{\textcolor{red}{{\footnotesize \marginnote{#3}}\sout{#1}\red{{#2}}}}

\providecommand{\ReviewL}[3]{\textcolor{red}{\reversemarginpar{\footnotesize\marginnote{#3}}\sout{#1}\red{{#2}}}}

\newcommand{\blue}[1]{{\color{blue} #1}}
\newcommand{\pmsize}[1]{\fontsize{6}{7}\selectfont#1}

\newcommand{\Ram}[1]{{\textnormal{\color{WildStrawberry} \textbf{Ram}: #1}}}
\newcommand{\jon}[1]{{\textnormal{\color{RoyalBlue} \textbf{Jon}: #1}}}
\newcommand{\yj}[1]{{\textnormal{\color{Purple} \textbf{Yong2Jon}: #1}}}
\newcommand{\yong}[1]{{\textnormal{\color{PineGreen} \textbf{Yong}: #1}}}

\newcommand{\addRam}[1]{{\textnormal{\color{WildStrawberry} #1}}}
\newcommand{\addyong}[1]{{\textnormal{{\color{PineGreen} #1}}}}
\newcommand{\addjon}[1]{{\textnormal{\color{RoyalBlue} #1}}}

\providecommand{\Review}[2]{\textcolor{cyan}{\sout{#1}{#2}}}
\newcommand{\new}[1]{{\color{Plum}{#1}}}
\newcommand{\stkout}[1]{{\color{Plum}\ifmmode\text{\sout{\ensuremath{#1}}}\else\sout{#1}\fi}}

\newcommand{\shrug}{\texttt{\raisebox{0.75em}{\char`\_}\char`\\\char`\_\kern-0.5ex(\kern-0.25ex\raisebox{0.25ex}{\rotatebox{45}{\raisebox{-.75ex}"\kern-1.5ex\rotatebox{-90})}}\kern-0.5ex)\kern-0.5ex\char`\_/\raisebox{0.75em}{\char`\_}}}

\providecommand{\methodname}{\text{CROWS}\xspace}
\providecommand{\methodnameNoGrad}{\text{CROWS(-)}\xspace}
\providecommand{\sparrows}{\text{SPARROWS}\xspace}
\providecommand{\armtd}{\text{ARMTD}\xspace}
\providecommand{\mpot}{\text{MPOT}\xspace}
\providecommand{\curobo}{\text{cuRobo}\xspace}
\providecommand{\chomp}{\text{CHOMP}\xspace}
\providecommand{\rdf}{\text{RDF}\xspace}
\providecommand{\trajopt}{\text{TrajOpt}\xspace}
\providecommand{\ipopt}{\text{IPOPT}\xspace}

\newcommand{\norm}[1]{\left\Vert#1\right\Vert}
\newcommand{\abs}[1]{\left\vert#1\right\vert}

\newtheorem{defn}{Definition}
\newtheorem{rem}[defn]{Remark}
\newtheorem{lem}[defn]{Lemma}
\newtheorem{prop}[defn]{Proposition}
\newtheorem{assum}[defn]{Assumption}
\newtheorem{ex}[defn]{Example}
\newtheorem{runx}{Running Example}
\newtheorem{thm}[defn]{Theorem}
\newtheorem{cor}[defn]{Corollary}
\newcommand{\regtext}[1]{\mathrm{\textnormal{#1}}}

\newcommand{\ol}[1]{\overline{#1}}
\newcommand{\ul}[1]{\underline{#1}}
\newcommand{\defemph}[1]{\emph{#1}}
\newcommand{\ts}[1]{\textsuperscript{#1}}
\newcommand{\nd}{n_d}
\newcommand{\relu}{\texttt{ReLU}}

\providecommand{\R}{\ensuremath \mathbb{R}}
\providecommand{\N}{\ensuremath \mathbb{N}}

\DeclarePairedDelimiter\ceil{\lceil}{\rceil}
\DeclarePairedDelimiter\floor{\lfloor}{\rfloor}

\providecommand{\Int}{\texttt{int}}
\providecommand{\nan}{\texttt{NaN}\xspace}
\providecommand{\INT}{\texttt{INT}}
\providecommand{\Sum}{\texttt{sum}}
\providecommand{\nX}{\texttt{nX}}
\providecommand{\Zaug}{\mathcal Z^{\text{aug}}}
\providecommand{\Zvel}{\mathcal Z^{\text{vel}}}
\providecommand{\zaug}{z^{\text{aug}}}
\providecommand{\zaugp}{z^{\text{aug}+}}
\providecommand{\zhi}{z^{\text{hi}}}
\providecommand{\zlo}{z^{\text{lo}}}
\providecommand{\zvel}{z^{\text{vel}}}
\providecommand{\zpos}{z^{\text{pos}}}
\providecommand{\cvel}{c^{\text{vel}}}
\providecommand{\Gvel}{G^{\text{vel}}}
\providecommand{\dzaug}{\dot z^{\text{aug}}}
\providecommand{\dzhi}{\dot z^{\text{hi}}}
\providecommand{\dzlo}{\dot z^{\text{lo}}}
\providecommand{\vlo}{v^{\text{lo}}}
\providecommand{\rlo}{r^{\text{lo}}}
\providecommand{\rw}{r_{\text{w}}}
\providecommand{\whlspd}{\omega_{\text{i}}}
\providecommand{\cus}{C_{\text{us}}}
\providecommand{\diag}{\texttt{diag}}
\providecommand{\slice}{\texttt{slice}}
\providecommand{\rot}{\texttt{rot}}
\providecommand{\ROT}{\texttt{ROT}}
\providecommand{\cost}{\texttt{cost}}
\providecommand{\Oego}{\mathcal O^\text{ego}}
\providecommand{\vegomax}{\nu^\text{ego}}
\providecommand{\vobsmax}{\nu^\text{obs}}
\providecommand{\tz}{t_0}
\providecommand{\tplan}{t_p}
\providecommand{\tnb}{t_\text{m}}
\providecommand{\tf}{t_\text{f}}
\providecommand{\tm}{t_\text{m}}
\providecommand{\tb}{t_\text{brake}}
\providecommand{\tstop}{t_\text{stop}}
\providecommand{\tfstop}{t_\text{fstop}}
\providecommand{\tsmall}{t_\text{small}}
\providecommand{\udes}{u^\text{des}}
\providecommand{\dudes}{\dot u^\text{des}}
\providecommand{\ubrk}{u^\text{brake}}
\providecommand{\usmall}{u^\text{small}}
\providecommand{\eu}{e_u}
\providecommand{\deu}{\dot e_u}
\providecommand{\boff}{b_u^\text{off}}
\providecommand{\bpro}{b_u^\text{pro}}
\providecommand{\hdes}{h^\text{des}}
\providecommand{\rdes}{r^\text{des}}
\providecommand{\drdes}{\dot r^\text{des}}
\providecommand{\duexp}{\dot u^\text{exp}}
\providecommand{\drexp}{\dot r^\text{exp}}
\providecommand{\fhi}{f^{\text{hi}}}
\providecommand{\flo}{f^{\text{lo}}}
\providecommand{\amax}{a^{\text{dec}}}
\providecommand{\Zaugini}{\Z_0^\text{aug}}
\providecommand{\caugini}{c_0^\text{aug}}
\providecommand{\Gaugini}{G_0^\text{aug}}
\providecommand{\hmid}{h^\text{mid}}
\providecommand{\hrad}{h^\text{rad}}
\providecommand{\uc}{u^\text{cri}}
\providecommand{\lambdac}{\lambda^\text{cri}}
\providecommand{\alphac}{\alpha^\text{cri}}
\providecommand{\pilo}{\pi_{1:4}}
\providecommand{\opt}{\texttt{(Opt)}}
\providecommand{\crowsopt}{\texttt{(CROWS-Opt)}}
\providecommand{\optref}{\hyperref[eq:optcost]{\opt{}}}
\providecommand{\crowsoptref}{\hyperref[eq:crowsoptcost]{\crowsopt{}}}
\providecommand{\bopt}{\texttt{(Batched Opt)} }
\providecommand{\nnopt}{\texttt{(NN-Opt)} }
\providecommand{\optth}{(\texttt{Opt}_{\texttt{TH}})}

\providecommand{\world}{W}
\providecommand{\workspace}{W_s}
\newcommand{\Aobs}{A_O}
\newcommand{\bobs}{b_O}
\newcommand{\hobs}{h\lbl{obs}}
\newcommand{\nObs}{n}
\newcommand{\NObs}{ N_{\mathscr{O}} }
\newcommand{\obsset}{\mathscr{O}}

\newcommand{\nhj}{n_{h,j}}
\newcommand{\Nhj}{ N_{h,j} }

\newcommand{\nq}{n_q}
\newcommand{\ns}{n_s}
\newcommand{\nt}{n_t}
\newcommand{\nf}{n_f}
\newcommand{\Nq}{ N_q }
\newcommand{\Ns}{ N_s }
\newcommand{\Nt}{ N_t }

\newcommand{\algorithmicbreak}{\textbf{break}}
\newcommand{\BREAK}{\STATE \algorithmicbreak}

\providecommand{\Psparse}{\mathcal{P}^\text{sparse}}
\providecommand{\Pdense}{\mathcal{P}^\text{dense}}
\providecommand{\PP}{\mathcal{P}}
\providecommand{\LL}{\mathcal{L}}
\providecommand{\W}{\mathcal{W}}
\providecommand{\D}{\mathcal{D}}
\providecommand{\X}{\mathcal{X}}
\providecommand{\K}{\mathcal{K}}
\providecommand{\Z}{\mathcal{Z}}
\providecommand{\XY}{\mathcal{XY}}
\renewcommand{\P}{\mathcal{P}}
\providecommand{\G}{\mathcal{G}}
\providecommand{\B}{\mathcal{B}}
\providecommand{\Z}{\mathcal{Z}}
\providecommand{\A}{\mathcal{A}}
\providecommand{\V}{\mathcal{V}}
\providecommand{\U}{\mathcal{U}}
\providecommand{\T}{\mathcal{T}}
\providecommand{\Y}{\mathcal{Y}}
\providecommand{\RR}{\mathcal{R}}
\providecommand{\Q}{\mathcal{Q}}
\providecommand{\HH}{\mathcal{H}}
\providecommand{\I}{\mathcal{I}}
\providecommand{\J}{\mathcal{J}}
\providecommand{\E}{\mathcal{E}}
\providecommand{\F}{\mathcal{F}}
\providecommand{\FO}{\regtext{\small{FO}}}
\providecommand{\FOt}{\mathcal{E}(t)}
\providecommand{\FOT}{\mathcal{E}(T,z_0,p)}
\providecommand{\FOz}{\xi}
\providecommand{\FOzj}{\xi_j}
\renewcommand{\SS}{\mathcal{S}}
\providecommand{\W}{\mathcal{W}}
\providecommand{\OO}{\mathcal{O}}

\newcommand{\setop}[1]{{\mathrm{\textnormal{\texttt{#1}}}}}
\newcommand{\numop}[1]{{\mathrm{\textnormal{\texttt{#1}}}}}

\newcommand{\qlim}{q_{j,\regtext{lim}}}
\newcommand{\dqlim}{\dot{q}_{j,\regtext{lim}}}
\newcommand{\ddqlim}{\ddot{q}_{j,\regtext{lim}}}
\newcommand{\ulim}{u_{j,\regtext{lim}}}
\newcommand{\timestep}{\Delta t}

\providecommand{\FRS}{FRS}
\providecommand{\FRSz}{\ensuremath FRS_\text{hip}}
\providecommand{\FRShipL}{\ensuremath FRS_{\text{\normalfont hip},L}}
\providecommand{\FRShipR}{\ensuremath FRS_{\text{\normalfont hip},R}}
\providecommand{\FRSxy}{\ensuremath FRS_\text{\normalfont pelvis}}
\providecommand{\zhip}{\ensuremath z_\text{hip}}
\providecommand{\zhipdot}{\ensuremath \dot{z}_\text{hip}}
\providecommand{\Tplan}{\ensuremath \tau_\text{plan}}
\providecommand{\Tcommand}{\ensuremath T_e}
\providecommand{\Tslot}{\ensuremath T_s}
\providecommand{\Tprocess}{\ensuremath T_c}
\providecommand{\Tsense}{\ensuremath T\sense}
\providecommand{\Tonline}{\ensuremath T_\text{plan}}
\providecommand{\Trrt}{\ensuremath T_\text{route}}
\providecommand{\Bl}{\ensuremath \underline{B}}
\providecommand{\Bu}{\ensuremath \overline{B}}
\providecommand{\FRSzOpt}{\textup{FRShipOpt}}
\providecommand{\FRSxyOpt}{\textup{FRSxyOpt}}
\providecommand{\Opt}{\textup{\texttt{Opt}}}
\newcommand{\obs}{_\text{obs}}
\newcommand{\plan}{_\text{plan}}
\newcommand{\sense}{_\text{sense}}
\newcommand{\hip}{_\text{hip}}
\newcommand{\pred}{_\text{pred}}
\newcommand{\goal}{_\text{goal}}
\providecommand{\hipzL}{_{\text{\normalfont hip},zL}}
\providecommand{\hipzR}{_{\text{\normalfont hip},zR}}
\providecommand{\hip}{_\text{\normalfont hip}}
\providecommand{\db}{\delta_b}
\providecommand{\ds}{\delta_s}
\providecommand{\M}{\mathcal{M}}
\providecommand{\totalParamNum}{8 }

\newcommand{\trunc}{^\text{trunc}}
\newcommand{\propa}{^\text{prop}}
\newcommand{\conv}{co}
\newcommand{\BV}{\text{BV}}

\providecommand{\x}{\mathbf{x}}
\renewcommand{\L}{\mathcal{L}}
\providecommand{\parammap}{\ensuremath M}
\providecommand{\buffer}{\textup{\texttt{buffer}}}
\providecommand{\sample}{\textup{\texttt{sample}}}
\providecommand{\disc}{\texttt{disc}}

\providecommand{\zthreshold}{0.75}

\providecommand{\BRTD}{\text{BipedRTD}}
\newcommand{\hi}{_\text{hi}}
\providecommand{\tfin}{t_\text{f}}
\providecommand{\T}{\ensuremath T}
\providecommand{\Oobs}{\mathcal{O}^{\text{obs}}_i}
\newcommand{\zonocg}[2]{ \text{\textless} #1,\; #2 \text{\textgreater}}

\newcommand{\des}{_\text{des}}
\newcommand{\pk}{_\text{peak}}

\newcommand{\kscale}{\eta_1}
\newcommand{\kjscale}{\eta_{j, 1}}
\newcommand{\koffset}{\eta_2}
\newcommand{\kjoffset}{\eta_{j, 2}}
\newcommand{\kvar}{x_k}
\newcommand{\kjvar}{x_{k_j}}
\newcommand{\kj}{k_j}
\newcommand{\Kj}{K_j}

\newcommand{\dist}{\mathbf{d}}
\newcommand{\sdf}{\mathbf{s_d}}
\newcommand{\SDF}{\texttt{SDF}}
\newcommand{\proj}{\phi}

\newcommand{\Nj}{\mathcal{J}}

\newcommand{\asdf}{\Tilde{s}}

\newcommand{\QP}{Alg. \ref{alg:rdf}\xspace}

\newcommand{\vi}[1]{v_{i}^{#1}}

\newcommand{\emptyarr}{[\ ]}
\newcommand{\zeros}{\textit{0}}
\newcommand{\ones}{\textit{1}}
\newcommand{\eye}{\regtext{\textit{I}}}

\newcommand{\cj}{c_j}
\newcommand{\cz}{c_z}
\newcommand{\co}{c_o}
\newcommand{\coj}{c_{o,j}}
\newcommand{\Gz}{G_z}
\newcommand{\Go}{G_o}
\newcommand{\Goj}{G_{o,j}}

\newcommand{\Oz}{\mathcal{O}_z}
\newcommand{\Oj}{\mathcal{O}_j}
\newcommand{\Ojz}{\mathcal{O}_{j,z}}

\newcommand{\Obs}{\mathcal{O}}

\newcommand{\Obsi}{O_i}
\newcommand{\Obsj}{O_j}
\newcommand{\Obsk}{O_k}
\newcommand{\Obst}{O(t)}
\newcommand{\ObsT}{O(T)}
\newcommand{\Obsit}{O_i(t)}
\newcommand{\Obsjt}{O_j(t)}
\newcommand{\Obskt}{O_k(t)}
\newcommand{\ObskT}{O_k(T)}
\newcommand{\Obsz}{\mathcal{O}}
\newcommand{\Obszi}{\mathcal{O}_i}
\newcommand{\Obszj}{\mathcal{O}_j}
\newcommand{\Obszk}{\mathcal{O}_k}
\newcommand{\Obszjk}{\mathcal{O}_{j,k}}
\providecommand{\Oobs}{\mathcal{O}^{\text{obs}}_i}

\providecommand{\sx}{{w_{\text{x}}}}
\providecommand{\sy}{{w_{\text{y}}}}

\newcommand{\pow}[1]{\mathcal{P}\!\left(#1\right)}

\providecommand{\sjo}{\mathcal{SJO}}
\providecommand{\slo}{\mathcal{SLO}}
\providecommand{\sfo}{\mathcal{SFO}}

\newcommand{\SJO}{{Spherical Joint Occupancy}\xspace}
\newcommand{\SLO}{{Spherical Link Occupancy}\xspace}
\newcommand{\SFO}{{Spherical Forward Occupancy}\xspace}

\newcommand{\pjone}{p_{j,1}}
\newcommand{\pjtwo}{p_{j,2}}
\newcommand{\pjonek}{p_{j,1}(k)}
\newcommand{\pjtwok}{p_{j,2}(k)}
\newcommand{\cjm}{c_{j,m}}
\newcommand{\cjmk}{c_{j,m}(k)}
\newcommand{\rjm}{r_{j,m}}
\newcommand{\rjmk}{r_{j,m}(k)}
\newcommand{\ljm}{\ell_{j,m}}
\newcommand{\ljmk}{\ell_{j,m}(k)}
\newcommand{\sk}{s(k)}
\newcommand{\sprimek}{s'(k)}

\newcommand{\interval}[1]{[ #1 ]}
\newcommand{\iv}[1]{[ #1 ]}
\newcommand{\nom}[1]{#1}
\newcommand{\pz}[1]{\mathbf{#1}}
\newcommand{\pzgreek}[1]{\bm{#1}}
\newcommand{\PZ}[1]{\mathcal{PZ}\left(#1\right)}


\newcommand{\pzk}[1]{\pz{ #1 ; k }}
\newcommand{\pzi}[1]{\pz{ #1 }(\pz{T_i};\pz{K})}
\newcommand{\pzij}[1]{\pz{ #1 }(\pz{T_i};\pz{K_j})}
\newcommand{\pzki}[1]{\pz{ #1 }(\pz{T_i};k)}
\newcommand{\pzjki}[1]{\pz{ #1 }_j (\pz{T_i};k )}
\newcommand{\pzjKi}[1]{\pz{ #1 }_j (\pz{T_i};K )}

\newcommand{\ith}{$i$\ts{th}}
\newcommand{\jth}{$j$\ts{th}}

\newcommand{\pzqi}{\pzi{q}}
\newcommand{\pzqli}{\pzi{q_l}}
\newcommand{\pzqdi}{\pzi{\dot{q}}}
\newcommand{\pzqdai}{\pzi{\dot{q}_{a}}}
\newcommand{\pzqddi}{\pzi{\ddot{q}}}
\newcommand{\pzqddai}{\pzi{\ddot{q}_{a}}}
\newcommand{\pzqdesi}{\pzi{q_{d}}}
\newcommand{\pzqddesi}{\pzi{\dot{q}_{d}}}
\newcommand{\pzqdddesi}{\pzi{\ddot{q}_{d}}}
\newcommand{\pzqdeski}{\pzki{q_{d}}}
\newcommand{\pzqddeski}{\pzki{\dot{q}_{d}}}
\newcommand{\pzqdddeski}{\pzki{\ddot{q}_{d}}}
\newcommand{\pzui}{\pzki{u}}
\newcommand{\pzqki}{\pzki{q}}
\newcommand{\pzqdki}{\pzki{\dot{q}}}
\newcommand{\pzuki}{\pzki{u}}

\newcommand{\pzqji}{\pzi{q_j}}
\newcommand{\pzqdji}{\pzi{\dot{q}_j}}
\newcommand{\pzqdaji}{\pzi{\dot{q}_{a,j}}}
\newcommand{\pzqddji}{\pzi{\ddot{q}_j}}
\newcommand{\pzqddaji}{\pzi{\ddot{q}_{a,j}}}
\newcommand{\pzqdesji}{\pzi{q_{d,j}}}
\newcommand{\pzqddesji}{\pzi{\dot{q}_{d,j}}}
\newcommand{\pzqdddesji}{\pzi{\ddot{q}_{d,j}}}
\newcommand{\pzqdesjki}{\pzki{q_{d,j}}}
\newcommand{\pzqddesjki}{\pzki{\dot{q}_{d,j}}}
\newcommand{\pzqdddesjki}{\pzki{\ddot{q}_{d,j}}}
\newcommand{\pzuji}{\pzki{u_j}}
\newcommand{\pzqjki}{\pzki{q_j}}
\newcommand{\pzqdjki}{\pzki{\dot{q}_j}}
\newcommand{\pzujki}{\pzki{u_j}}

\newcommand{\pzujKi}{\pz{u}(\pzqAi, \nomparams, \intparams)}
\newcommand{\pzFKjki}{\pz{FK_j}(\pzqki)}
\newcommand{\pzFOjki}{\pz{FO_j}(\pzqki)}
\newcommand{\pzFOki}{\pz{FO}(\pzqki)}
\newcommand{\pzFOjkibuf}{\pz{FO_{j}^{buf}}(\pzqki)}
\newcommand{\pzFKjKi}{\pz{FK_j}(\pzqi)}
\newcommand{\pzFOjKi}{\pz{FO_j}(\pzqi)}
\newcommand{\Pj}{\mathcal{P}_j}
\newcommand{\Hjh}{\mathcal{H}_j^{(h)}}
\newcommand{\Ajh}{A_j^{(h)}}
\newcommand{\bjh}{b_j^{(h)}}

\newcommand{\pzg}{g}
\newcommand{\pzv}{x}
\newcommand{\pze}{\alpha}
\newcommand{\pzn}{{n_g}}
\newcommand{\pzng}{{n_g}}
\newcommand{\pzgi}{g_i}
\newcommand{\pznG}{{n_G}}
\newcommand{\pzGi}{G_i}

\newcommand{\pzei}{\alpha_i}

\newcommand{\tvar}{x_t}
\newcommand{\tvari}{x_{t_{i}}}

\newcommand{\q}{q(t)}
\newcommand{\qd}{\dot{q}(t)}
\newcommand{\qdd}{\ddot{q}(t)}
\newcommand{\qa}{q_a(t)}
\newcommand{\qadot}{\dot{q}_a(t)}
\newcommand{\qaddot}{\ddot{q}_a(t)}
\newcommand{\qak}{q_a(t; k)}
\newcommand{\qakdot}{\dot{q}_a(t; k)}
\newcommand{\qakddot}{\ddot{q}_a(t; k)}
\newcommand{\qdes}{q_d(t)}
\newcommand{\qdesdot}{\dot{q}_d(t)}
\newcommand{\qdesddot}{\ddot{q}_d(t)}
\newcommand{\qdesk}{q_d(t; k)}
\newcommand{\qdeskdot}{\dot{q}_d(t; k)}
\newcommand{\qdeskddot}{\ddot{q}_d(t; k)}


\newcommand{\qstart}{q_{\text{start}}}
\newcommand{\qgoal}{q_{\text{goal}}}
\newcommand{\qj}{q_j(t)}
\newcommand{\ql}{q_l(t)}
\newcommand{\qdj}{\dot{q}_{j}(t)}
\newcommand{\qddj}{\ddot{q}_{j}(t)}
\newcommand{\qkj}{q_j(t; k)}
\newcommand{\qdkj}{\dot{q}_{j}(t; k)}
\newcommand{\qddkj}{\ddot{q}_{j}(t; k)}
\newcommand{\qaj}{q_{a, j}(t)}
\newcommand{\qajdot}{\dot{q}_{a, j}(t)}
\newcommand{\qajddot}{\ddot{q}_{a, j}(t)}
\newcommand{\qdesj}{q_{d, j}(t)}
\newcommand{\qdesjdot}{\dot{q}_{d, j}(t)}
\newcommand{\qdesjddot}{\ddot{q}_{d, j}(t)}
\newcommand{\qdeskj}{q_{d, j}(t; k)}
\newcommand{\qdeskjdot}{\dot{q}_{d, j}(t; k)}
\newcommand{\qdeskjddot}{\ddot{q}_{d, j}(t; k)}

\newcommand{\homtrans}{H}

\newcommand{\FK}{\regtext{\small{FK}}}
\newcommand{\IK}{\regtext{\small{IK}}}
\newcommand{\IO}{\regtext{\small{IO}}}
\newcommand{\FS}{\regtext{\small{FS}}}
\newcommand{\IS}{\regtext{\small{IS}}}
\newcommand{\FC}{\regtext{\small{FC}}}
\newcommand{\IC}{\regtext{\small{IC}}}
\newcommand{\ID}{\regtext{\small{ID}}}

\newcommand{\pzCjiK}{\pz{C_j}(\pz{T_i}; \pz{K})}
\newcommand{\pzCjik}{\pz{C_j}(\pz{T_i}; k)}
\newcommand{\pzCjpik}{\pz{C_{j+1}}(\pz{T_i}; k)}
\newcommand{\pzuj}{\pz{u_j}(\pz{T_i})}
\newcommand{\pzrj}{\pz{r_j}(\pz{T_i})}

\newcommand{\Sjq}{S_j(\q)}
\newcommand{\Sjqk}{S_j(q(t;k))}
\newcommand{\Sjpq}{S_{j+1}(\q)}
\newcommand{\Sjpqk}{S_{j+1}(q(t;k))}

\newcommand{\Sjik}{S_{j}(\pzqki)}

\newcommand{\SjiK}{{S_{j}(\pzi{q})}}

\newcommand{\Sjpik}{S_{j+1}(\pzqki)}
\newcommand{\SjpiK}{S_{j+1}(\pzi{q})}

\newcommand{\sjik}{s_{j,i}(k)}
\newcommand{\spjik}{s'_{j,i}(k)}

\newcommand{\Sbarjimk}{{\bar{S}_{j,i,m}(\pzqki)}}

\newcommand{\TCjqk}{TC_{j}(q(t;k))}
\newcommand{\TCjik}{TC_{j}(\pzqki)}
\newcommand{\TCjiK}{TC_{j}(\pzi{q})}

\providecommand{\slok}{\mathcal{SLO}(\pzqki)}
\providecommand{\sfok}{\mathcal{SFO}(\pzqki)}
\providecommand{\sfojk}{\mathcal{SFO}_j(\pzqki)}

\providecommand{\Bji}{\mathcal{B}_{j,i}}
\providecommand{\Bjidelta}{\hat{\mathcal{B}}_{j,i}(\delta_j)}
\providecommand{\BjiDelta}{\hat{\mathcal{B}}_{j,i}(\Delta_j)}

\maketitle
\thispagestyle{plain}
\pagestyle{plain} 

\begin{abstract}
Safe motion planning algorithms are necessary for deploying autonomous robots in unstructured environments. 
Motion plans must be safe to ensure that the robot does not harm humans or damage any nearby objects.
Generating these motion plans in real-time is also important to ensure that the robot can adapt to sudden changes in its environment.
Many trajectory optimization methods introduce heuristics that balance safety and real-time performance, potentially increasing the risk of the robot colliding with its environment.
This paper addresses this challenge by proposing Conformalized Reachable Sets for Obstacle Avoidance With Spheres (\methodname).
\methodname is a novel real-time, receding-horizon trajectory planner that generates probablistically-safe motion plans. 
Offline, \methodname learns a novel neural network-based representation of a sphere-based reachable set that overapproximates the swept volume of the robot's motion.
\methodname then uses conformal prediction to compute a confidence bound that provides a probabilistic safety guarantee on the learned reachable set.
At runtime, \methodname performs trajectory optimization to select a trajectory that is probabilstically-guaranteed to be collision-free.
We demonstrate that \methodname outperforms a variety of state-of-the-art methods in solving challenging motion planning tasks in cluttered environments while remaining collision-free.
Code, data, and video demonstrations can be found at \url{https://roahmlab.github.io/crows/}.
\end{abstract}

\section{Introduction}\label{sec:intro}
Robot manipulators have the potential to transform multiple facets of how humans live and work.
This includes using robots to complete daily chores in peoples' homes, replacing humans in performing difficult, dangerous, or repetitive tasks at construction sites, and assisting medical professionals with life-saving surgeries.
In each of these settings, it is essential that the robot remain safe at all times to prevent colliding with obstacles, damaging high-value objects, or harming nearby humans.
It is also critical that the robot generates motion plans in real-time to ensure that any given task is performed efficiently or the robot can quickly adjust its behavior to react to local changes in its environment.

A modern approach to motion planning typically involves combining a high-level planner, a mid-level trajectory planner, and a low-level tracking controller into a hierarchical framework \cite{holmes2020armtd, michaux2023armour}.
The high-level planner generates a path between the robot's start and goal configurations consisting of a sequence of discrete waypoints. 
The mid-level trajectory planner computes time-dependent positions, velocities, and accelerations that move the robot from one waypoint to the next. 
The low-level tracking controller generates control inputs that attempt to minimize deviations between the robot's actual motion and the desired trajectory. 
For example, one may use a sampling-based planner such as an RRT$^{\ast}$ to generate a set of discrete waypoints for a trajectory optimization algorithm such as \trajopt \cite{Schulman2014trajopt}, and track the resulting trajectories with an inverse dynamics controller \cite{Caccavale2020}.
\begin{figure}[t]
    \centering
    \includegraphics[width=0.9\columnwidth]{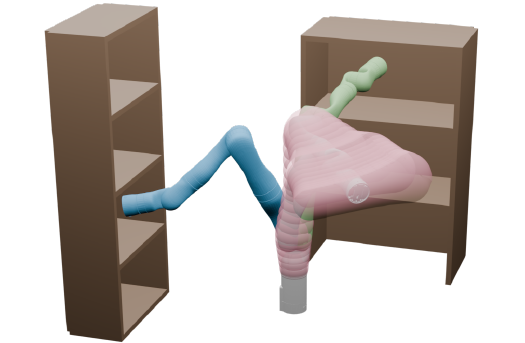}
    \caption{%
    This paper presents \methodname, a method that generates probabilistically-safe motion plans in cluttered environments.
    Prior to planning, \methodname learns neural safety representation with confidence bounds computed by conformal prediction.
    \methodname also constructs an exact signed distance function of the obstacles in the scene (Assum. \ref{assum:obstacles}).
    At runtime, \methodname combines the signed distance function with the conformalized safety representation (pink) to generate probabilistically-safe trajectories between the start (blue) and goal (green) in a receding horizon manner.
    Each trajectory is selected by solving a nonlinear optimization problem with the learned safety representation (pink) guaranteeing with high probability that the robot remains collision-free.}
    \label{fig:network}
    \vspace*{-0.5cm}
\end{figure}
\begin{figure*}[t]
    \centering
    \includegraphics[width=\textwidth]{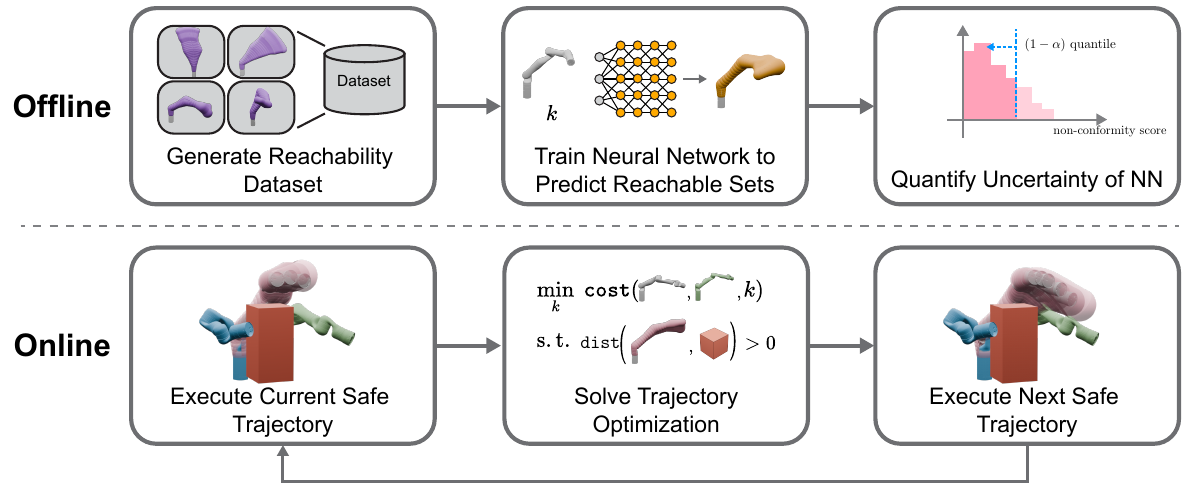}
    \caption{
    Overview of \methodname{}, a probabilistically-safe receding-horizon trajectory planner.
    Offline, \methodname constructs a dataset to train a network that predicts spherical overapproximations of parameterized swept volumes of a robot arm. 
    After training, conformal prediction is used to quantify the uncertainty of the network's predictions.
    Online, the output of the network is buffered by a nonconformity score and used as a safety constraint during trajectory optimization. 
    Finally, \methodname{} executes the trajectory while generating a plan for the next iteration. 
    }
    \label{fig:overview_figure}
    \vspace*{-0.5cm}
\end{figure*}

While variations of this planning framework have been demonstrated to work on various robotic platforms \cite{kousik2017safe, holmes2020armtd, michaux2023armour}, several limitations prevent this approach from being widely deployed in the real-world. 
For instance, this approach can become computationally expensive as the complexity of the environment or robot increases, making it less practical for applications requiring both real-time performance and safety guarantees.
Many algorithms also introduce heuristics to improve computation speed such as reducing the number of collision checks to achieve real-time performance; however, this may come at the expense of robot safety.  
Furthermore, many algorithms assume that the robot's dynamics are fully known, while in reality there can be considerable model uncertainty. 
Unfortunately, both of these factors may increase the potential for the robot to collide with obstacles.

To address this challenge, this paper proposes Conformalized Reachable Sets for Obstacle Avoidance With Spheres (\methodname), a neural network-based safety representation that can be efficiently integrated into a mid-level trajectory optimization algorithm (Fig. \ref{fig:overview_figure}).
\methodname extends \cite{michaux2024sparrows} by learning an overapproximation of the swept volume (\emph{i.e.} reachable set) of a serial robot manipulator that is composed entirely of spheres.
Prior to planning, a neural network is trained to approximate the sphere-based reachable set (Sec. \ref{subsec:networks}).
Then, \methodname applies conformal prediction to compute a confidence bound that provides a probabilistic safety guarantee (Sec. \ref{subsec:conformal_prediction}).
Finally, \methodname uses the conformalized reachable set and its learned gradient to solve an optimization problem to generate probabilistically-safe trajectories online (Sec. \ref{subsec:planning}).
In Sec. \ref{subsec:motion_planning_experiments}, we demonstrate that this novel reachable set formulation enables planning in extremely cluttered environments.




\subsection{Related Work}\label{sec:related_work}
Safe motion planning algorithms ensure that a robot can move from one configuration to another in its environment while avoiding collisions for all time.
Ideally, one would compute the robot's swept volume \cite{blackmore_differential_1990, blackmore_analysis_1992, blackmore_analysis_1994, blackmore_sweep-envelope_1997, blackmore_trimming_1999, spatial_planningPerez} and ensure that it does not intersect with any obstacles nor enter any unwanted regions of its workspace.
Alternatively, many algorithms approximate the swept volume using occupancy grids, convex polyhedra, or CAD models \cite{Campen2010PolygonalBE, Kim2003FastSV, Gaschler2013RobotTA}.
However, both approaches are often intractable or suffer from high computational costs while approximate swept volume algorithms may be overly conservative \cite{Gaschler2013RobotTA, ekenna2015}.
This may limit their utility to offline motion planning \cite{perrin2012svfootstep}.

An alternative approach to robot collision avoidance involves modeling the robot or the environment with simple collision primitives such as spheres \cite{duenser2018manipulation, gaertner2021collisionfree}, ellipsoids \cite{brito2020model}, capsules \cite{dube2013humanoids,khoury2013humanoids}, and then performing collision-checking along a given trajectory at discrete time instances.
This is common for state-of-the-art trajectory optimization-based approaches \cite{Zucker2013chomp, Schulman2014trajopt, le2023mpot, sundaralingam2023curobo}.
However, the resulting trajectories cannot be considered safe as collision avoidance is not enforced for all time.

Reachability-based Trajectory Design (RTD) \cite{kousik2017safe}, a recent approach to real-time motion planning, uses (polynomial) zonotopes \cite{kochdumper2020polyzono} to construct reachable sets that overapproximate all possible robot positions corresponding to a pre-specified family of parameterized trajectories.
Notably, RTD's reachable sets are constructed to ensure that its obstacle-avoidance constraints are satisfied in continuous-time.
\sparrows \cite{michaux2024sparrows}, a recent extension to RTD, utilizes sphere-based reachable sets to generate certifiably-safe motion plans that are considerably less conservative than previous approaches \cite{holmes2020armtd, michaux2023rdf, michaux2023armour, brei2024waitr}.
The purpose of this paper is to introduce a neural representation of \sparrows with uncertainty quantification for real-time \emph{probabilistically-safe} motion planning.

Several methods have been developed to quantify the uncertainty of neural network models such as Monte Carlo dropout \cite{gal2016dropout, shamsi2021improving, lutjens2019safe}, Laplace approximation \cite{daxberger2021laplace, ritter2018scalable, goli2024bayes}, and deep ensembles \cite{chua2018deep, lakshminarayanan2017simple}.
More recently, conformal prediction \cite{angelopoulos2021gentle, shafer2008tutorial} has gained popularity due to its ability to provide probabilistic guarantees on coverage.
Conformal prediction has been employed in various robotics applications including: ensuring the safety of learning-based object detection systems \cite{dixit2024perceive}; quantifying the uncertainty of human inputs during teleoperation \cite{zhao2024conformalized}; ensuring safety in environments with dynamic obstacles \cite{lindemann2023safe,lekeufack2024conformal}; and applying conformal prediction to align uncertainty in large language models, enabling them to decide when to seek human assistance.
Notably, \cite{lin2024verification} is closely related to our work, as it enhances the efficiency of reachable set computations for high-dimensional systems with neural network while verifying it with conformal prediction.

\section{Background}\label{sec:background}
\begin{figure}[t]
    \centering
    \includegraphics[width=0.93\columnwidth]{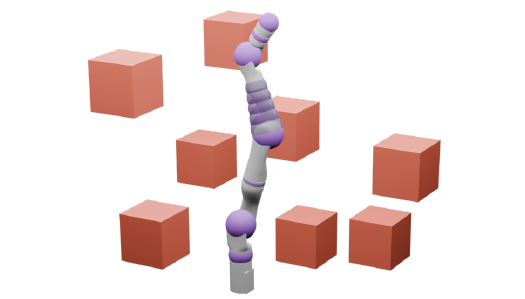}
    \caption{A visualization of the robot arm (grey), the obstacles (red), and a subset of the Spherical Forward Occupancy (purple) over a single time step $T_i$.
    The volume of each joint (solid purple) and link (transparent purple) is overapproximated by a collection of spheres  (Thm. \ref{thm:sparrows}).
    }
    \label{fig:arm_env}
    \vspace*{-0.5cm}
\end{figure}
This section summarizes the background necessary for the development of \methodname in Sec. \ref{sec:approach}.

\subsection{Mathematical Preliminaries}
Sets, subspaces, and matrices are typeset using capital letters.
Let $\R$ and $\N$ denote the spaces of real numbers and natural numbers, respectively.
Subscripts are primarily used as an index or as a label for relevant countable sets.
For example, if $n_{\alpha} \in \N$, then we denote the set $N_{\alpha} = \{1, \cdots, n_{\alpha} \}$.
Given a set $\mathcal A$, denote its power set as $P(\mathcal A)$.
Given a set $\Omega \subset \mathbb{R}^{\nd}$, $\conv(\Omega)$ denote its convex hull.

\subsection{Arm Occupancy}
\label{app:modeling_arm_occupancy}
This subsection describes how to overapproximate the forward occupancy, or swept volume, of a moving robot arm.
Consider a compact time interval $T \subset \R$.
We define a trajectory for the robot's configuration as $q: T \to Q \subset \R^{\nq}$ and a trajectory for the velocity as $\dot{q}: T \to \R^{\nq}$.
To facilitate the later discussion of the robot's occupancy, we begin by restating an assumption \cite[Ass. 4]{michaux2024sparrows} about the robot model:
\begin{assum}\label{assum:robot}
The robot operates in an three-dimensional workspace, denoted $\workspace \subset \R^{3}$.
There exists a reference frame called the base frame, denoted the $0$\ts{th} frame, that indicates the origin of the robot's kinematic chain within the workspace.
The robot is fully actuated and composed of only revolute joints, where the $j$\ts{th} joint actuates the robot's $j$\ts{th} link.
The robot's $j$\ts{th} joint has position and velocity limits given by $\qj \in [\qlim^-, \qlim^+]$ and $\qdj \in [\dqlim^-, \dqlim^+]$ for all $t \in T$, respectively. 
\end{assum}

Let $L_j \subset \workspace \subset \R^3$ denote the volume occupied by the robot's $j$\ts{th} link with respect to the $j$\ts{th} reference frame. 
Then the \emph{forward occupancy} of the $j$\ts{th} link is the map $\FO_j: Q \to \pow{\workspace}$ defined as
\begin{align}\label{eq:link_occupancy}
     \FO_j(\q) &= p_j(\q) \oplus R_j(\q) L_j,
\end{align}
where $p_j(\q)$ and $R_j(\q)$ are computed by the forward kinematics \cite{spong2005textbook} and specify the pose of the $j$\ts{th} joint, and $R_j(\q) L_j$ is the rotated volume of link $j$.
The volume occupied by the entire arm in the workspace is the union of the link volumes defined by the map $\FO: Q \to \workspace$:
\begin{align}\label{eq:forward_occupancy}
    \FO(\q) = \bigcup_{j = 1}^{\nq} \FO_j(\q) \subset \workspace. 
\end{align}
Due to the potentially complex geometry of a robot's links, we reiterate an assumption \cite[Ass. 5]{michaux2024sparrows} that facilitates the construction of an overapproximation of the forward occupancy:
\begin{assum}\label{assum:joint_link_occupancy}
    Given a robot configuration $\q$ and any $j \in \{1,\dots, \nq\}$, there exists a ball with center $p_j(\q)$ and radius $r_j$  that overapproximates the volume occupied by the $j$\ts{th} joint in $\workspace$.
    We further assume that link volume $L_j$ is a subset of the tapered capsule formed by the convex hull of the balls overapproximating the $j$\ts{th} and $j+1$\ts{th} joints.
\end{assum}

Following Assum. \ref{assum:joint_link_occupancy}, we now define the ball $\Sjq$ overapproximating the volume occupied by the $j$\ts{th} joint as 
\begin{equation}\label{eq:joint_occupancy}
    \Sjq = \mathcal{B}(p_j(\q), r_j)
\end{equation}
and the tapered capsule $TC_j(q(t))$ overapproximating the $j$\ts{th} link as
\begin{equation}
 TC_j(q(t)) = \conv\Big( \Sjq \cup \Sjpq \Big).
\end{equation}
Then, the volumes occupied by the $j$\ts{th} link \eqref{eq:link_occupancy} and the entire arm \eqref{eq:forward_occupancy} can be overapproximated by
\begin{equation}\label{eq:link_capsule_occupancy}
     \FO_j(\q)  \subset TC_j(q(t))
\end{equation}
and
\begin{equation} \label{eq:forward_capsule_occupancy}
    \FO(\q) \subset \bigcup_{j = 1}^{\nq} TC_j(q(t)) \subset \workspace, 
\end{equation}
respectively.

Note that if the forward occupancy (or \emph{reachable set}) $\FO(q(T))$ does not intersect with the environment, then the arm is \emph{collision-free} over the time interval $T$. 
To facilitate the exposition of our approach, we summarize the construction of the robot's safety representation by restating \cite{michaux2024sparrows}[Thm. 10]:
\begin{thm}\label{thm:sparrows}
Given a serial manipulator with $n_q \in \N$ revolute joints, a time partition $T$ of a finite set of intervals, $T_i$ (i.e., $T = \cup_{i=1}^{n_t} T_i$), the swept volume corresponding to the robot's motion over $T$ is overapproximated by a collection of $L_2$ balls in $\R^3$, which we call the \SFO (SFO \cite{michaux2024sparrows} defined as
\begin{equation}\label{eq:simple_sfo}
\sfo = \cup_{j=1}^{n_q} \cup_{i=1}^{n_t} \cup_{m=1}^{n_S} S_{j,i,m}(q(T_i; k)),
\end{equation}
where each $S_{j,i,m}(q(T_i; k))$ is an $L_2$ ball in $\mathbb{R}^3$, $n_S \in \N$ is a parameter that specifies the number of closed balls overapproximating each of the robot's links, and $k$ is a trajectory parameter that characterizes the motion of the robot over $T$.
\end{thm}
\noindent Note that one can explicitly construct an $\sfo$ that satisfies this assumption using the approach described in \cite{michaux2024sparrows}.

\subsection{Environment Modeling}\label{subsec:sdf}
The $\sfo$ constructed in Thm. \ref{thm:sparrows} is composed entirely of spheres, where each sphere is a collision primitive consisting of a point with a safety margin corresponding to its radius.
For simplicity, we assume that each obstacle $\mathcal{O}$ is a polytope.
Therefore, to make use of this sphere-based representation for trajectory planning, we state the result of \cite{michaux2024sparrows} [Lem. 11] as an assumption describing the existence of an obstacle exact signed distance:
\begin{assum}[Environment Signed Distance Function]
\label{assum:obstacles}
Given the $\sfo$ and a convex obstacle polytope $\mathcal{O} \in \R^3$, there exists a known function, $\sdf(\sfo, \mathcal{O})$, that computes the exact signed distance between $\sfo$ and $\mathcal{O}$.
\end{assum}

\section{Trajectory Optimization Formulation}\label{sec:approach}
The method described in \cite{michaux2024sparrows} computes provably-safe motion plans by solving a nonlinear optimization program in a receding-horizon manner:
\begin{align}
    \label{eq:optcost}
    &\underset{k \in K}{\min} &&\numop{cost}(q_{\text{goal}}, k) \hspace{3cm} \opt&  \\
    \label{eq:optpos}
    &&& \hspace*{-0.75cm} q_j(T_i; k) \subseteq [\qlim^-, \qlim^+]  \hspace{1cm} \forall (i,j) \in N_t \times N_q \\
    \label{eq:optvel}
    &&& \hspace*{-0.75cm}\dot{q}_j(T_i; k) \subseteq [\dqlim^-, \dqlim^+] \hspace{1cm} \forall (i,j) \in N_t \times N_q \\
    \label{eq:optcolcon}
    &&& \hspace*{-0.75cm}\sdf(\sfo, \mathcal{O}_{n}) > 0 \hspace{1.95cm} \forall n \in N_{\mathcal{O}}, 
\end{align}
where $\mathcal{O}_{n}$ is the $n$\ts{th} obstacle for $n \in N_{\mathcal{O}}$.
Offline, we pre-specify a continuum of trajectories over a compact set $K \subset \R^{n_k}$ such that $n_k \in \N$.
Then each trajectory, $q(t;k)$, is defined over a compact time interval $T$ and is uniquely determined by a \textit{trajectory parameter}  $k \in K$.
The cost function \eqref{eq:optcost} ensures the robot moves towards a user- or task-defined goal $q_{\text{goal}}$.
The constraints \eqref{eq:optpos}--\eqref{eq:optvel} ensure that the trajectory remains feasible and does not violate the robot's joint position and velocity limits, respectively.
The last constraint \eqref{eq:optcolcon} guarantees safety by ensuring that the robot's forward occupancy does not collide with any obstacles in the environment.

In the remainder of this section, we describe how \methodname replaces \eqref{eq:optcolcon} with a fast, neural representation with probabilistic safety-guarantees.

\subsection{Neural Network Models}
\label{subsec:networks}
In this subsection, we describe how \methodname learns neural network representations of the centers and radii of $\sfo$  in \eqref{eq:simple_sfo} and their gradients with respect to the trajectory parameter.
\begin{figure}[t]
    \centering
    \includegraphics[width=0.93\columnwidth]{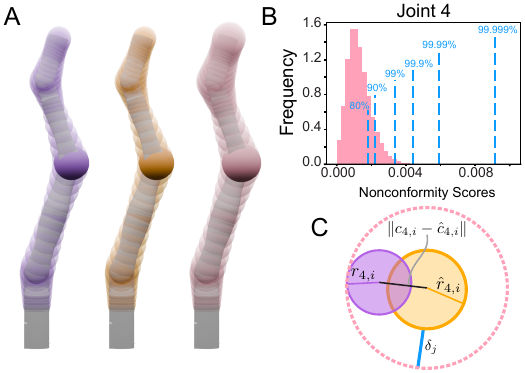}
    \caption{
    A visual illustration of the construction of \methodname's conformalized reachable sets.
    Panel (A) compares the ground truth (left, purple), predicted (middle, orange), and conformalized (right, pink) reachable sets.
    To construct the conformalized reachable set, \methodname first defines the nonconformity score in \eqref{eq:buffer}, which is the minimum buffer (blue) to ensure that the predicted sphere (orange) encloses the ground truth sphere (purple) (Panel C).
    The distribution of the nonconformity scores for joint $4$ over the interval $T_i$ with the values defining the quantiles indicated in blue (Panel B).
    Next, conformal prediction computes a confidence bound that upper bounds the nonconformity scores with a probability of $1-\epsilon$.
    The predicted joint sphere (A, middle) is then expanded by the size of this confidence bound to give the conformalized joint sphere (A, right).
    Finally, applying this procedure for all of the joint spheres gives the conformalized neural (A, right) reachable set that is guaranteed to cover the ground truth reachable set with probability greater than $(1-\epsilon)^{n_q +1}$ (Thm. \ref{thm:crows}).
    }
    \label{fig:conformal}
    \vspace*{-0.5cm}
\end{figure}

\subsubsection{Neural SFO}
\label{subsec:neural_sfo}
Given a family of parameterized trajectories $q(t;k)$ for $k \in K$, we train a neural network to predict the centers $c_{j,i}(q_j(T_i; k))$ and radii $r_{j,i}(q_j(T_i; k))$ for all $(i,j) \in N_t \times N_q$ of the Spherical Forward Occupancy.
For simplicity, we have dropped the subscript $m$ to indicate that the center and radii networks only predict the spheres that overapproximate the joint volumes.

Following training, we use \cite{michaux2024sparrows}[Lem. 9, Thm. 10] to construct the remaining spheres that overapproximate the links.
The network takes as input $x = (q_0, \dot{q}_0, k, i)$, which consists of a concatenation of vectors corresponding to the initial joint positions $q_0$, initial joint velocities $\dot{q}_0$, trajectory parameter $k$, and the $i$\ts{th} time index.
The output of the network is $y = (\hat{c}_{j,i}, \hat{r}_{j,i}) \in \mathbb{R}^{4(n_q + 1)}$.
We discuss specific training details in Sec. \ref{subsubsec:network_training}.

\subsubsection{Neural SFO Gradient} \label{subsec:neural_sfo_gradient}
To facilitate real-time solutions of \optref, we also train a neural network to output $\widehat{\frac{\partial c_{j,i}}{\partial k}}$, which is a prediction of the gradient of each joint center with respect to the trajectory parameter $k$.
Because the radii of $\sfo$ do not depend on $k$, we do not train its corresponding gradient network.

\subsection{Conformal Prediction}
\label{subsec:conformal_prediction}
Next, we describe the use of conformal prediction to formulate probabilistic guarantees for collision avoidance.
\subsubsection{Calibration} 
We first construct a calibration set $\mathcal{D}_{cal}=\{(x_d,y_d)\}_{d=1}^{N_{cal}}$ with i.i.d. samples, where $N_{cal} \in \N$. 
Let $\Bji = \mathcal{B}(c_{j,i}, r_{j,i})$ be a ground truth sphere and $\Bjidelta = \mathcal{B}(\hat{c}_{j,i}, \hat{r}_{j,i} + \delta_{j})$ be the corresponding prediction.
Note that $\Bji$ corresponds to $S_{j,i}(q(T_i; k))$ in Thm. \ref{thm:sparrows}.
Then for the calibration step, we define a nonconformity score $\delta_{j}$ for each joint sphere such that
\begin{align} \label{eq:buffer}
    &\delta_{j}=\max(||c_{j,i}-\hat{c}_{j,i}||_2+r_{j,i}-\hat{r}_{j,i}, \;0)
\end{align}
is minimum buffer required for $\mathcal{B}_{j,i} \subseteq \hat{\mathcal{B}}_{j,i}(\delta_j)$.

Given the nonconformity scores over the calibration set $ \mathcal{D}_{cal}$, the nonconformity score $\delta_{j,d'}$ on a new sample $(x_{d'},y_{d'})$  follows a gaurantee conditioned on the calibration set $ \mathcal{D}_{cal}$ with the probability of $1-\rho$ \cite{vovk2012conditional, dixit2024perceive}:
\begin{align}\label{eq:dataset-conditioned}
&\mathbb{P}(\delta_{j,d'}\leq \Delta_{j,1- \hat{\epsilon}}| \mathcal{D}_{cal} )\geq \text{Beta}_{N_{cal}+1-\nu,\nu}(\rho)
\end{align}
where $\nu:=\lfloor (N_{cal}+1)\hat{\epsilon}\rfloor$, $\text{Beta}_{N_{cal}+1-\nu,\nu}(\rho)$ is the $\rho$-quantile of the Beta distribution, $\hat{\epsilon}$ is a user-defined variable for desired coverage, and $\Delta_{j,1- \hat{\epsilon}}$ is the $\lceil(N_{cal}+1)(1-\hat{\epsilon})\rceil$-th quantile of the nonconformity scores  over the calibration set $ \mathcal{D}_{cal}$.
To facilitate the following exposition, let 
\begin{align}
\mathbb{P}(\cdot|\mathcal{D}_{cal}) &= \mathbb{P}(\cdot), \\
\Delta_{j,1- \hat{\epsilon}} &= \Delta_j,
\end{align}
and
\begin{equation}
\text{Beta}_{N_{cal}+1-v,v}(\rho) = 1-\epsilon. 
\end{equation}
Then the guarantee in \eqref{eq:dataset-conditioned} can be written as:
\begin{align}
\label{eq:cp-gaurantee} 
&\mathbb{P}(\delta_{j} \leq \Delta_{j} )\geq 1-\epsilon
\end{align}
where $1-\epsilon$ is desired coveraged specified by $\hat{\epsilon}$.

\subsubsection{Conformalized Reachable Sets}
Given \eqref{eq:cp-gaurantee}, the conformalized sphere $\hat{\mathcal{B}}_{j,i}(\Delta_j)$ is guaranteed to enclose the ground truth sphere $\mathcal{B}_{j,i}$ with the following probability:
\begin{align}
&\mathbb{P}(\mathcal{B}_{j,i} \subseteq \BjiDelta)\\
&\geq \mathbb{P}(\mathcal{B}_{j,i}\subseteq \Bjidelta)\cdot\mathbb{P}(\Bjidelta \subseteq \BjiDelta) \\
&= \mathbb{P}(\Bjidelta \subseteq \BjiDelta) \\
&= \mathbb{P}(\delta_{j}\leq \Delta_{j} ) \geq 1-\epsilon.
\end{align}

This approach can be extended to all joint occupancy spheres to construct a probabilistic safety guarantee to ensure that the ground truth reachable set is collision-free with probability greater than $(1-\epsilon)^{{n_q+1}}$ over the $i$\ts{th} time interval $T_i$.
We summarize this result in the following theorem whose proof can be found in App. \ref{appendix:conformal_proof}:
\begin{thm}\label{thm:crows}
Given $\mathbb{P}(\mathcal{B}_{j,i} \subseteq \BjiDelta)\geq 1-\epsilon$ for $j\in N_q$, the following probability holds under the condition $\sdf(\Hat{\sfo}_{i},\obsset)>0$:
\begin{align}&\mathbb{P}(\sdf(\text{FO}_i,\obsset)>0) \geq (1-\epsilon)^{{n_q+1}}\end{align}
where $\text{FO}_i$ denotes the forward occupancy over a single time interval $T_i$ (\eqref{eq:forward_occupancy}), $\Hat{\sfo}_i$ is the \SFO constructed from $\BjiDelta$ for all $j \in N_q$ over the same time interval $T_i$ by Theorem \ref{thm:sparrows}, $\obsset$ represents a set of obstacles defined as $\{\mathcal{O}_n \mid n \in N_{\mathcal{O}}\}$, and $\sdf(\cdot, \obsset) = \min_{n \in N_{\mathcal{O}}} \sdf(\cdot, \mathcal{O}_n)$.
\end{thm}

\subsection{Online Trajectory Optimization}
\label{subsec:planning}
After training, we generate models to predict the spheres of the $\sfo$. 
Using this representation, we can reformulate the optimization problem described by \eqref{eq:optcost}--\eqref{eq:optcolcon} into:
\begin{align}
    \label{eq:crowsoptcost}
    &\underset{k \in K}{\min} &&\numop{cost}(q_{\text{goal}}, k) \hspace{1cm} \crowsopt&  \\
    \label{eq:crowsoptpos}
    &&& \hspace*{-0.75cm} q_j(T_i; k) \subseteq [\qlim^-, \qlim^+]  \hspace{1cm} \forall (i,j) \in N_t \times N_q \\
    \label{eq:crowsoptvel}
    &&& \hspace*{-0.75cm}\dot{q}_j(T_i; k) \subseteq [\dqlim^-, \dqlim^+] \hspace{1cm} \forall (i,j) \in N_t \times N_q \\
    \label{eq:crowsoptcolcon}
    &&& \hspace*{-0.75cm}\sdf(\Hat{\sfo}_{i},\obsset)>0 \hspace{2cm} \forall i \in N_t
\end{align}
where $\Hat{\sfo}$ is the conformalized reachable set in Thm. \ref{thm:crows}.
\section{Results}
\begin{table}[t]
        \centering
            \begin{tabular}{ | c || c | c | c ||}
                \hline 
                Methods & \multicolumn{3}{c||}{mean constraint eval. time [ms]} \\\hline
                \# Obstacles (s) & 10 & 20 & 40 \\\hline \hline
                \methodname  & 3.5 ± 0.2 & 4.1 ± 0.1 & 5.5 ± 0.1 \\ \hline
                \methodnameNoGrad & 6.3 ± 1.5 & 6.9 ± 1.2 & 8.3 ± 1.2 \\ \hline
                \sparrows & \textbf{3.1± 0.1} & \textbf{3.8 ± 0.1 }& \textbf{5.2 ± 0.1}\\ \hline
                \armtd  & 4.1 ± 0.3 & 5.4 ± 0.5 & 8.1 ± 0.7 \\ \hline
                \end{tabular}
        \caption{
        Mean runtime for constraint and constraint gradient evaluation across 100 \textbf{Random Obstacle Scenarios} under a $0.5$s planning time limit $\downarrow$.}
        \vspace*{-0.5cm}
        \label{tab:constraint_evaluations}
\end{table}
\begin{table}[t]
        \centering
            \begin{tabular}{ | c || c | c | c ||}
                \hline 
                Methods & \multicolumn{3}{c||}{mean planning time [s]} \\\hline
                \# Obstacles (s) & 10 & 20 & 40 \\\hline \hline
                \methodname & \textbf{0.14  ± 0.29} & \textbf{0.16 ± 0.10} & \textbf{0.21 ± 0.10} \\ \hline
                \methodnameNoGrad  & 0.17 ± 0.10 & 0.20 ± 0.10 & 0.24 ± 0.11 \\ \hline
                \sparrows      & 0.16 ± 0.08  & 0.18 ± 0.08 & 0.24 ± 0.09 \\ \hline
                \armtd           & 0.24 ± 0.09 & 0.38 ± 0.10 & \textcolor{red}{0.51 ± 0.04} \\ \hline
                \end{tabular}
        \caption{
        Mean per-step planning time across 100 \textbf{Random Obstacle Scenarios} under a 0.5s planning time limit $\downarrow$. 
        \textcolor{red}{Red} indicates that the average planning time limit has been exceeded.}
        \vspace*{-1cm}
        \label{tab:planning_time}
\end{table}

\begin{table}[t]
    \centering
    \begin{tabular}{|c||c||c||c||c||}
    \hline
    Methods & \multicolumn{4}{c||}{\# Successes}\\\hline
    Environment & \multicolumn{3}{c||}{Random Obstacle} & Realistic\\\hline
    \# Obstacles & 10 & 20 & 40 & N.A.\\
    \hline \hline
    \methodname & \textbf{87} & 58 &37 &13\\\hline
    \methodnameNoGrad & 83 & 60 & 30 &12\\\hline
    \sparrows  & \textbf{87} & \textbf{62} & \textbf{40} & \textbf{14}\\\hline
    \armtd & 56 & 17 & 0 &8\\\hline
    \chomp \cite{Zucker2013chomp} & 76  & 40  & 15& 10 \\
    \hline
    \trajopt \cite{Schulman2014trajopt} & 33 (\textcolor{red}{67}) & 9 (\textcolor{red}{91}) & 6
    (\textcolor{red}{94}) & 5 (\textcolor{red}{9})\\
    \hline
    \mpot \cite{le2023mpot} & 58 (\textcolor{red}{42}) & 23 (\textcolor{red}{77}) & 9 (\textcolor{red}{91}) & 6 (\textcolor{red}{8})\\\hline
    \curobo \cite{sundaralingam2023curobo} &  59 (\textcolor{red}{41}) & 45 (\textcolor{red}{55}) & 22 (\textcolor{red}{78}) & 5 (\textcolor{red}{9})\\\hline
    \end{tabular}
    \caption{
    Number of successes out of 100 trials on \textbf{Random Obstacle Scenarios} (the first three columns) and out of 14 trials on the \textbf{Realistic Scenarios} (the last column).   
    \textcolor{red}{Red} indicates the number of failures due to collision.}
    \label{tab:random_obstacles_0.5s}
    \vspace*{-1cm}
\end{table}

We demonstrate the performance of \methodname in both simulation and hardware experiments.

\subsection{Implementation Details}
A computer with 12 Intel(R) Core(TM) i7-8700K CPU @ 3.70GHz and an NVIDIA RTX A6000 GPU was used for the motion planning experiment in Sec. \ref{subsec:motion_planning_experiments}.
The \methodname{} model is built and trained with Pytorch \cite{paszke2017automatic}. 
The trajectory optimization \crowsoptref was solved with \ipopt \cite{ipopt-cite}. 

\subsubsection{Simulation and Simulation Environment}
\label{subsubsec:simulation}
We simulate the Kinova Gen3 7-DOF serial manipulator \cite{kinova-user-guide}.
The robot's collision geometry is provided as a mesh, which we utilize with the trimesh library \cite{trimesh} to check for collisions with obstacles. 
For simplicity, we do not check self-collisions of the robot and all obstacles are static, axis-aligned cubes.
We assume that the start and goal configurations of the robot are collision-free. 

\subsubsection{Training Details}
\label{subsubsec:network_training}
Fully-connected networks were trained to predict the centers, radii, and gradients of the centers with respect to the trajectory parameter.
These networks have a width 1024 with 3, 9, and 12 hidden layers, respectively.
The radii network uses the ReLU activation function while the others use GELU \cite{hendrycks2023gelu}.
As discussed in \ref{subsec:neural_sfo}, the networks only predict the centers and radii that overapproximate the joint volumes.
The networks also do not predict the center and radius of the base joint sphere as it remains constant for all time.
For the loss function, we used the mean squared error between the target and the prediction.
Each network was trained using the AdamW \cite{kingma2017adam} optimizer with learning rate 0.0003, beta (0.9,0.999), and weight decay 0.0001.
The training, validation, and calibration datasets consisted of 8e5, 1e5, and 1e5 samples, respectively.


\subsection{Model Prediction Accuracy}
\label{subsec:prediction_accuracy}
The mean and maximum errors of the center predictions across all joints are $0.87$cm and $2.10$cm, respectively. 
For the radius predictions, the mean and maximum errors are $0.12$ cm and $0.59$cm. 
The median relative error of the neural SFO gradient is $1.25\%$.
The nonconformity scores \eqref{eq:buffer} are computed using the calibration set.
The 99.9\ts{th} percentile of the nonconformity scores for joints $2$ through $8$ are ($0.05$, $0.18$, $0.35$, $0.44$, $0.56$, $0.64$, $0.75$) cm, measured from the proximal to distal joint.
In the trajectory optimization \crowsoptref, an uncertainty bound of $\hat{\epsilon} = 0.001$ was used. 

\subsection{Runtime Comparisons}
\label{subsec:runtime}
We compare the runtime performance of \methodname to \sparrows \cite{michaux2024sparrows} and \armtd \cite{holmes2020armtd} under a planning time limit while varying the number of obstacles.
We also include a comparison to \methodnameNoGrad, which uses PyTorch's built-in \texttt{autograd} function to compute the gradient rather than the learned model introduced in Sec. \ref{subsec:neural_sfo_gradient}.
We measure the mean constraint evaluation time as well as the mean planning time.
Constraint evaluation includes the time to compute the constraints and the constraint gradients.
Planning time includes the time required to construct the reachable sets and the total time required to solve \optref{} including constraint gradients evaluations.
For each case, the results are averaged over 100 trials.

Tables \ref{tab:constraint_evaluations} and \ref{tab:planning_time} summarize the runtime comparisons under a planning time limit of $0.5s$.
\sparrows has the lowest mean constraint evaluation time followed by \methodname, \methodnameNoGrad, and \armtd.
\methodname, on the other hand, has the lowest per-step planning time followed by \sparrows, \methodnameNoGrad, and \armtd.
These results indicate that learned gradient improves the solve time of \crowsoptref when using the learned safety representation.

\subsection{Motion Planning Experiments}
\label{subsec:motion_planning_experiments}
We compare the performance of \methodname to \sparrows, \armtd, \chomp, \trajopt, \mpot, and \curobo on two sets of planning tasks.
The first set of tasks, \textbf{Random Obstacle Scenarios}, contains $n_{\mathcal{O}}=10$, $20$, or $40$ axis-aligned boxes that are $20$cm on each side.
For each number of obstacles, we generate 100 scenes with obstacles placed randomly such that the start and goal configurations are collision-free.
The second set of tasks, \textbf{Realistic Scenarios}, contains 14 scenes with geometric features similar to those found in a household setting. 
For both planning tasks, a failure occurs if \methodname, \sparrows, or \armtd fails to find a plan for two consecutive planning iterations or if a collision occurs.
\methodname, \sparrows, and \armtd are given a planning time limit of $0.5$s and a maximum of 150 planning iterations to reach the goal. 
\chomp{}, \trajopt{}, and \curobo{} are given planning time limited of 100s, 30s, and 2s, respectively.

\begin{figure}[t]
    \centering
    \includegraphics[width=\columnwidth]{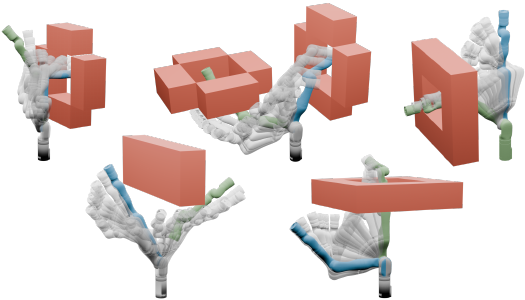}
    \caption{
    A subset of \textbf{Realistic Scenarios} where \methodname succeeds.
    The start, goal, and intermediate poses are shown in blue, green, and grey (transparent), respectively. 
    Obstacles are shown in red (transparent).
    }
    \vspace*{-0.5cm}
    \label{fig:realistic_scenarios}
\end{figure}

Tables \ref{tab:random_obstacles_0.5s} compare the success rate of \methodname to the baselines on the \textbf{Random Obstacle Scenarios} and \textbf{Realistic Scenarios} (Fig. \ref{fig:realistic_scenarios}), respectively.
\sparrows achieves the highest success rate on both sets of tasks followed by \methodname or \methodnameNoGrad.
Note that both \methodname and \methodnameNoGrad are competitive with \sparrows while also remaining collision-free.
In contrast, the baselines \trajopt, \mpot, and \curobo all experience collisions.

\subsection{Hardware Experiments}
\label{subsec:hardware_experiments}
We demonstrate \methodname solving motion planning tasks on a real hardware setup using various initial and goal configurations. 
The hardware setup follows a similar architecture to \cite{michaux2024splanning}, where a high-level planner and mid-level trajectory planner run in parallel, and a low-level controller tracks trajectories using the Recursive Newton-Euler Algorithm (RNEA) \cite{luh1980line} alongside a PD controller. 
In the real-world experiments, \methodname successfully performs real-time trajectory planning, reaching the goal configuration without colliding with the environment.
Video demonstrations can be found at \url{https://roahmlab.github.io/crows/}.

\section{Conclusion and Future Directions}
We present \methodname, a method for generating real-time motion plans for robot manipulators in cluttered environments. 
We show \methodname remains safe across many motion planning trials and has performance comparable to \sparrows, a state-of-the-art, model-based trajectory optimizer.
We also demonstrate that learning a representation of $\sfo$'s gradient leads to improvements in both planning time and success rate.
Future directions will extend \methodname to higher-dimensional systems such as humanoid robots as well as learning conformalized representations of the robot's dynamics. 
Furthermore, we anticipate that embedding \methodname within end-to-end architectures for planning in more general scene representations is also a promising direction for future research.


\renewcommand{\bibfont}{\normalfont\footnotesize}
{\renewcommand{\markboth}[2]{}
\printbibliography}

\clearpage
\appendices
\section{Proof of Theorem 5}\label{appendix:conformal_proof}
\begin{proof}
The convex hull of two conformalized spheres encloses the convex hull of the corresponding ground truth spheres with probability:
\begin{equation}
\label{eq:tc-1-gaurantee}
\mathbb{P}(\text{TC}_{i,j} \subseteq \hat{\text{TC}}_{i,j})
\geq \prod_{j'=j}^{j+1}\mathbb{P}(\mathcal{B}_{j',i} \subseteq \hat{\mathcal{B}}_{j',i}(\Delta_{j'})) \geq (1-\epsilon)^2
\end{equation}
where $\text{TC}_{i,j}=\conv(\bigcup_{j' = j}^{j+1}(\mathcal{B}_{j',i}))$
and $\hat{\text{TC}}_{i,j}=\conv(\bigcup_{j' = j}^{j+1}(\hat{\mathcal{B}}_{j',i}(\Delta_{j'}))$.
This probability extends across all the joints as follows:
\begin{align} 
&\mathbb{P}(\bigcup_{j = 1}^{\nq} \text{TC}_{i,j} \subseteq \bigcup_{j = 1}^{\nq} \hat{\text{TC}}_{i,j}) \geq (1-\epsilon)^{{n_q+1}}\end{align}
Because $\text{FO}_i\subseteq\bigcup_{j = 1}^{\nq} \text{TC}_{i,j}$ and $\bigcup_{j = 1}^{\nq} \hat{\text{TC}}_{i,j} \subseteq \Hat{\sfo}_{i}$,
the following guarantee holds:
\begin{align}
&\mathbb{P}(\text{FO}_i\subseteq\Hat{\sfo}_{i}) \geq (1-\epsilon)^{{n_q+1}}\end{align}
Therefore, the probability that the signed distance between the ground truth forward occupancy $\text{FO}_i$ and obstacle set $\obsset$ remains positive is bounded by:
\begin{align}
&\mathbb{P}(\sdf(\text{FO}_i,\obsset)>0) \\
&= \mathbb{P}(\text{FO}_i\subseteq\Hat{\sfo}_{i}) \cdot \mathbb{P}(\sdf(\Hat{\sfo}_{j},\obsset)>0) \\
&\geq (1-\epsilon)^{{n_q+1}} \cdot \mathbb{P}(s(\Hat{\sfo}_{j},\obsset)>0)\end{align}
If we enforce $\mathbb{P}(\sdf(\Hat{\sfo}_{j},\obsset)>0)=1$, then the following probability holds:
\begin{align}
&\mathbb{P}(\sdf(\text{FO}_i,\obsset)>0)\geq(1-\epsilon)^{{n_q+1}}.
\end{align} 
\end{proof}

\end{document}